\documentclass[12pt]{article}

\usepackage{amsmath}
\usepackage{amsfonts}

\usepackage{graphicx}

\usepackage{hyperref}

\usepackage{cleveref}

\usepackage{enumitem}

\usepackage{geometry}

\usepackage{setspace}

\usepackage{fancyhdr}

\usepackage{titlesec}
\usepackage{titletoc}

\usepackage{textcomp}

\usepackage{biblatex}

\title{Enhancing Trust in LLMs: Algorithms for Comparing and Interpreting LLMs}
\author{Nik Bear Brown \\ Northeastern University and Bear Brown \& Company}
\date{}

\begin{document}
\maketitle

\begin{abstract}
This paper presents a survey of evaluation techniques aimed at enhancing the trustworthiness and understanding of Large Language Models (LLMs). Amidst growing reliance on LLMs across various sectors, ensuring their reliability, fairness, and transparency has become paramount. We explore a range of algorithmic methods and metrics designed to assess LLMs' performance, identify weaknesses, and guide their development towards more trustworthy and effective applications. Key evaluation metrics discussed include Perplexity Measurement, Natural Language Processing (NLP) evaluation metrics (BLEU, ROUGE, METEOR, BERTScore, GLEU, Word Error Rate, and Character Error Rate), Zero-Shot Learning Performance, Few-Shot Learning Performance, Transfer Learning Evaluation, Adversarial Testing, and Fairness and Bias Evaluation. We also introduce innovative approaches such as LLMMaps for stratified evaluation, Benchmarking and Leaderboards for competitive assessment, Stratified Analysis for in-depth understanding, Visualization of Bloom’s Taxonomy for cognitive level accuracy distribution, Hallucination Score for quantifying inaccuracies, Knowledge Stratification Strategy for hierarchical analysis, and the use of Machine Learning Models for Hierarchy Generation. Furthermore, we highlight the indispensable role of Human Evaluation in capturing nuances that automated metrics may overlook. Together, these techniques form a robust framework for evaluating LLMs, aiming to enhance transparency, guide development, and align assessments with the goal of establishing user trust in these advanced language models. In future papers, we will describe the visualization of these metrics as well as demonstrate the use of each approach on practical examples.
\end{abstract}

\section{Introduction}
Evaluating Large Language Models (LLMs) is a nuanced process that extends beyond technical metrics to encompass considerations of social alignment, transparency, safety, and trustworthiness. Liu (2023) stresses the significance of ensuring LLMs align with human intentions, adhering to societal norms and regulations. Liao (2023) advocates for a human-centered transparency approach, focusing on the varied needs of all stakeholders involved. Huang (2023) delves into the safety and reliability of LLMs, suggesting the adoption of Verification and Validation (V\&V) techniques to mitigate risks and conduct thorough assessments. Karabacak (2023) highlights the unique challenges in the medical sector, calling for comprehensive strategies that include clinical validation, ethical considerations, and adherence to regulatory standards. Together, these perspectives emphasize the essential roles of transparency and trust in evaluating LLMs, especially for their application in real-world scenarios.

The evaluation of LLMs is fundamental to building trust and ensuring transparency in these advanced AI systems. As LLMs increasingly permeate various sectors such as education, healthcare, and legal advising, the importance of their careful assessment becomes paramount. This discussion explores the complexities of LLM evaluation, underscoring transparency and trust as pivotal elements for their successful integration and acceptance in society.

\section{The Imperative for Transparency}
Transparency in LLMs refers to the clarity and openness regarding how models are trained, how they operate, and how they make decisions. This transparency is pivotal for several reasons:
\begin{itemize}
    \item \textbf{Understanding Model Decisions}: Stakeholders, including users, developers, and regulators, must understand the basis of an LLM's outputs. Transparent models allow for the identification of the data and algorithms that drive decisions, facilitating insights into their reliability.
    \item \textbf{Detecting and Mitigating Biases}: Transparent evaluation processes enable the identification of biases in LLM outputs. By understanding how and why biases occur—whether due to training data or model architecture—developers can implement targeted interventions to mitigate them.
    \item \textbf{Facilitating Model Improvements}: A transparent evaluation framework helps pinpoint specific areas where LLMs excel or falter. This clarity is crucial for guiding ongoing model refinement and ensuring that improvements are aligned with ethical standards and societal needs.
    \item \textbf{Selecting the Right Model}: Transparency aids in choosing the best LLM for specific tasks by comparing models on performance, training, and ethical standards. This ensures compatibility with user needs and regulatory requirements.
    \item \textbf{Ensuring Compliance and Trust}: Transparent evaluations and decision-making processes help meet regulatory standards and build user trust, highlighting a commitment to ethical AI.
    \item \textbf{Promoting Collaborative Development}: Openness in model evaluation encourages shared problem-solving, leading to innovative solutions and model enhancements.
    \item \textbf{Supporting Lifelong Learning and Adaptation}: Transparent evaluation facilitates ongoing model monitoring and updates, keeping LLMs relevant and aligned with evolving standards and needs.
\end{itemize}

\section{The Quest for Trust}
Trust in LLMs hinges on their ability to perform tasks accurately, ethically, and reliably. Trustworthiness is built through establishing the right metrics. In this survey paper, we will focus on the following metrics:
\begin{itemize}
    \item \textbf{Perplexity Measurement}: Evaluates model fluency by measuring how well a model predicts a sample. While perplexity is a valuable metric, it's not without limitations. It primarily focuses on the probabilistic prediction of words without directly measuring semantic accuracy or coherence.
    \item \textbf{NLP evaluation metrics}: BLEU, ROUGE, METEOR, BERTScore, GLEU, Word Error Rate (WER), and Character Error Rate (CER). These metrics are used to assess various aspects of machine-generated text, such as translation quality, summarization effectiveness, semantic similarity, and transcription accuracy, in the context of natural language processing tasks. Each metric focuses on different elements of text generation and comprehension, providing a comprehensive framework for evaluating the performance of NLP models and systems.
    \item \textbf{Zero-Shot Learning Performance}: Assesses the model's ability to understand tasks without explicit training.
    \item \textbf{Few-Shot Learning Performance}: Evaluates how well a model performs tasks with minimal examples.
    \item \textbf{Transfer Learning Evaluation}: Tests the model's ability to apply learned knowledge to different but related tasks.
    \item \textbf{Adversarial Testing}: Identifies model vulnerabilities by evaluating performance against inputs designed to confuse or trick the model.
    \item \textbf{Fairness and Bias Evaluation}: Measures model outputs for biases and fairness across different demographics.
    \item \textbf{Robustness Evaluation}: Assesses model performance under varied or challenging conditions.
    \item \textbf{LLMMaps}: A novel visualization technique for stratified evaluation across subfields, emphasizing the identification of areas where LLMs excel or require improvement, particularly in reducing hallucinations.
    \item \textbf{Benchmarking and Leaderboards}: Common tools that involve LLMs answering questions from large Q\&A datasets to test their accuracy.
    \item \textbf{Stratified Analysis}: A detailed, stratified analysis across various knowledge subfields for a comprehensive understanding of LLMs' strengths and weaknesses.
    \item \textbf{Visualization of Bloom’s Taxonomy}: Displays accuracy for each level of Bloom's Taxonomy in a pyramidal fashion to understand the distribution of accuracy across different cognitive levels.
    \item \textbf{Hallucination Score}: A metric introduced within LLMMaps to quantify instances where the model provides inaccurate or unsupported responses.
    \item \textbf{Knowledge Stratification Strategy}: A top-down approach for creating a hierarchical knowledge structure within Q\&A datasets, enabling nuanced analysis and interpretation.
    \item \textbf{Utilization of Machine Learning Models for Hierarchy Generation}: Employing LLMs to generate and categorize each question into the most fitting subfield, based on overarching topics derived from the dataset.
    \item \textbf{Sensitivity Analysis}: This involves altering inputs slightly and observing the changes in the model's output. For LLMs, sensitivity analysis can reveal how changes in word choice or sentence structure affect the generated text, highlighting the model's responsiveness to specific linguistic features.
    \item \textbf{Feature Importance Methods}: These methods identify which parts of the input data are most influential in determining the model's output. In the context of LLMs, feature importance can show which words or phrases in a sentence contribute most significantly to the model's predictions or decisions.
    \item \textbf{Shapley Values}: Originating from cooperative game theory, Shapley values provide a way to distribute a "payout" (i.e., the output prediction) among the "players" (i.e., input features) based on their contribution. Applying Shapley values to LLMs can quantify the contribution of each word or token to the model's output, offering a fair and robust measure of feature importance.
    \item \textbf{Attention Visualization}: Many LLMs, especially those based on the Transformer architecture, use attention mechanisms to weigh the importance of different parts of the input data. Visualizing these attention weights can illustrate how the model focuses on specific parts of the input text when generating responses, revealing patterns in how it processes information.
    \item \textbf{Counterfactual Explanations}: This involves modifying parts of the input data to see how these changes alter the model's output, essentially asking "what if" questions. For LLMs, counterfactual explanations can help understand the conditions under which the model's decisions or predictions change, shedding light on its reasoning process.
    \item \textbf{Language-Based Explanations}: These are natural language explanations generated by the model itself or another model to explain the reasoning behind a given output. In LLMs, generating language-based explanations can make the model's decision-making process more accessible and interpretable to humans.
    \item \textbf{Embedding Space Analysis}: This technique explores the vector representations (embeddings) of words or phrases used by the model to understand semantic and syntactic relationships. Analyzing the embedding space of LLMs can reveal how the model organizes and relates concepts, offering insights into its understanding of language.
    \item \textbf{Computational Efficiency and Resource Utilization}: Peak Memory Consumption, Memory Bandwidth Utilization, CPU/GPU Utilization Percentage, FLOPS (Floating Point Operations Per Second), Inference Time, Number of Parameters, Model Storage Size, Compression Ratio, Watts per Inference/Training Hour, Parallelization Efficiency, Batch Processing Capability.
    \item \textbf{Human Evaluation}: Involves human judges assessing the quality, relevance, or coherence of model-generated text.
\end{itemize}

\section{Perplexity Measurement}
Perplexity Measurement serves as a fundamental metric in the evaluation of Language Models (LMs), including Large Language Models (LLMs), by quantifying their fluency and predictive capabilities. Sundareswara (2008) highlights its importance in assessing model fluency, emphasizing its role in measuring how effectively a model can predict a sequence of words. The methodology for perplexity estimation has seen various innovations; notably, Bimbot (1997, 2001) introduced a novel scheme based on a gambling approach and entropy bounds, offering an alternative perspective that enriches the metric's applicability. This approach was further validated through comparative evaluations, underscoring its relevance. Additionally, Golland (2003) proposed the use of permutation tests for estimating statistical significance in discriminative analysis, suggesting a potential avenue for applying statistical rigor to the evaluation of language models, including their perplexity assessments.

While perplexity is invaluable for gauging a model's fluency, it is not without its limitations. Its primary focus on the probabilistic prediction of words means that it does not directly measure semantic accuracy or coherence, areas that are crucial for the comprehensive evaluation of LMs, especially in complex applications. This metric, deeply rooted in information theory, remains a critical tool for understanding how well a probability model or distribution can anticipate a sample, providing essential insights into the model's understanding of language.

\subsection{Understanding Perplexity}
Perplexity is calculated as the exponentiated average negative log-likelihood of a sequence of words, given a language model. A lower perplexity score indicates a better performing model, as it suggests the model is more confident (assigns higher probability) in its predictions. Conversely, a higher perplexity score suggests the model is less certain about its predictions, equating to less fluency.

\subsection{Application in Evaluating LLMs}
\begin{itemize}
    \item \textbf{Model Comparison}: Perplexity allows researchers and developers to compare the performance of different LLMs on the same test datasets. It helps in determining which model has a better understanding of language syntax and structure, thereby predicting sequences more accurately.
    \item \textbf{Training Diagnostics}: During the training phase, perplexity is used as a diagnostic tool to monitor the model's learning progress. A decreasing perplexity trend over training epochs indicates that the model is improving in predicting the training data.
    \item \textbf{Model Tuning}: Perplexity can guide the hyperparameter tuning process by indicating how changes in model architecture or training parameters affect model fluency. For instance, adjusting the size of the model, learning rate, or the number of layers can have a significant impact on perplexity, helping developers optimize their models.
    \item \textbf{Domain Adaptation}: In scenarios where LLMs are adapted for specific domains (e.g., legal, medical, or technical fields), perplexity can help evaluate how well the adapted model performs in the new domain. A lower perplexity in the target domain indicates successful adaptation.
    \item \textbf{Language Coverage}: Perplexity can also shed light on the model's coverage and understanding of various languages, especially for multilingual models. It helps in identifying languages that the model performs well in and those that may require further data or tuning for improvement.
\end{itemize}

\subsection{Limitations}
While perplexity is a valuable metric, it's not without limitations. It primarily focuses on the probabilistic prediction of words without directly measuring semantic accuracy or coherence. Therefore, it's often used in conjunction with other evaluation metrics (like those mentioned earlier: BLEU, ROUGE, etc.) that can assess semantic similarity and relevance to provide a more holistic evaluation of LLMs.

In summary, perplexity is a foundational metric in NLP for evaluating the fluency and predictive accuracy of language models, playing a critical role in the development and refinement of LLMs.

\section{Natural Language Processing (NLP) Evaluation Metrics}
A range of NLP evaluation metrics, including BLEU, ROUGE, METEOR, BERTScore, GLEU, WER, and CER, are used to assess LLMs in various tasks (Blagec, 2022). However, these metrics have been found to have low correlation with human judgment and lack transferability to other tasks and languages. This raises concerns about the adequacy of these metrics in reflecting model performance (Blagec, 2022). Despite these limitations, LLMs have shown promise in radiology NLP, with some models demonstrating strengths in interpreting radiology reports (Liu, 2023). However, in domain-specific applications, such as Wikipedia-style survey generation, LLMs have exhibited shortcomings, including incomplete information and factual inaccuracies (Gao, 2023). Similarly, in medical evidence summarization, LLMs have been found to struggle with identifying salient information and generating factually inconsistent summaries (Tang, 2023). These studies highlight the need for more robust evaluation metrics and the importance of considering the limitations of existing ones.

\subsection{BLEU (Bilingual Evaluation Understudy)}
\textbf{Use:} Primarily for machine translation quality assessment.\\
\textbf{How It Works:} Compares machine-generated translations to one or more reference translations, focusing on the precision of n-grams (contiguous sequences of n items from a given sample of text).\\
\textbf{Strengths:} Simple, widely used, correlates well with human judgment at the corpus level.\\
\textbf{Limitations:} Lacks sensitivity to meaning preservation, grammatical correctness, and does not consider recall.

\subsection{ROUGE (Recall-Oriented Understudy for Gisting Evaluation)}
\textbf{Use:} Evaluates summarization quality, including both extractive and abstractive methods.\\
\textbf{How It Works:} Measures the overlap of n-grams, word sequences, and word pairs between the generated summaries and reference summaries, emphasizing recall.\\
\textbf{Strengths:} Captures content selection effectiveness, supports multiple reference summaries.\\
\textbf{Limitations:} May not fully represent the quality of summaries (e.g., coherence, readability).

\subsection{METEOR (Metric for Evaluation of Translation with Explicit ORdering)}
\textbf{Use:} Another metric for translation assessment that extends beyond BLEU's capabilities.\\
\textbf{How It Works:} Aligns generated text to reference texts considering exact matches, synonyms, stemming, and paraphrasing, with penalties for incorrect word order.\\
\textbf{Strengths:} Better correlation with human judgment on sentence-level evaluation, compensates for some of BLEU's shortcomings.\\
\textbf{Limitations:} More complex computation, potential for overfitting specific test sets.

\subsection{BERTScore}
\textbf{Use:} Evaluates semantic similarity between generated text and reference text.\\
\textbf{How It Works:} Utilizes contextual embeddings from models like BERT to compute similarity scores between words in generated and reference texts, aggregating these scores for an overall measurement.\\
\textbf{Strengths:} Captures deeper semantic meanings not evident in surface-level matches; robust to paraphrasing.\\
\textbf{Limitations:} Computationally intensive, interpretation of scores can be less intuitive.

\subsection{GLEU (Google BLEU)}
\textbf{Use:} Tailored for evaluating shorter texts, such as those in machine translation and language understanding tasks.\\
\textbf{How It Works:} Similar to BLEU but adapted to work better on shorter sentences, often used internally by Google.\\
\textbf{Strengths:} More sensitive to errors in short texts.\\
\textbf{Limitations:} Like BLEU, may not fully account for semantic accuracy.

\subsection{Word Error Rate (WER)}
\textbf{Use:} Commonly used in speech recognition to evaluate the accuracy of transcribed text.\\
\textbf{How It Works:} Compares the transcribed text with a reference text, calculating the proportion of errors (substitutions, deletions, insertions).\\
\textbf{Strengths:} Straightforward, intuitive metric for transcription accuracy.\\
\textbf{Limitations:} Does not account for semantic meaning or grammatical correctness.

\subsection{Character Error Rate (CER)}
\textbf{Use:} Similar to WER but evaluates transcription accuracy at the character level.\\
\textbf{How It Works:} Measures the minimum number of character insertions, deletions, and substitutions required to change the transcribed text into the reference text.\\
\textbf{Strengths:} Useful for languages where character-level evaluation is more indicative of transcription quality.\\
\textbf{Limitations:} Like WER, focuses on surface errors without accounting for semantic content.

\subsection{Application in LLM Evaluation}
In evaluating LLMs, these metrics are often used together to provide a multifaceted view of model performance across various tasks. For instance, while BLEU and METEOR might be used to evaluate translation models, ROUGE could be applied to summarization tasks, and BERTScore for tasks requiring semantic evaluation. WER and CER are particularly relevant for voice-driven applications where speech-to-text accuracy is critical.

\subsection{Challenges and Considerations}
No single metric can capture all aspects of language model performance. It's crucial to select metrics that align with the specific goals of the task at hand. Moreover, the interpretation of these metrics should consider their limitations and the context of their application. For comprehensive evaluation, combining these metrics with qualitative analysis and human judgment often yields the most insightful assessments of LLM capabilities.

\section{Zero-Shot Learning Performance}
Recent studies have shown that large language models (LLMs) like GPT-3 can achieve strong zero-shot learning performance, even without task-specific fine-tuning datasets (Brown, 2020). This is further supported by the work of Meng (2022), who demonstrated the potential of using both unidirectional and bidirectional PLMs for zero-shot learning of natural language understanding tasks. Puri (2019) also highlighted the use of natural language descriptions for zero-shot model adaptation, achieving significant improvements in classification accuracy. These findings collectively underscore the impressive zero-shot learning capabilities of LLMs, which are crucial for their generalization and adaptability to a wide range of tasks.

\subsection{Understanding Zero-Shot Learning Performance}
\textbf{Concept:} Zero-shot learning involves evaluating the model's performance on tasks it has not seen during its training phase. It relies on the model's pre-existing knowledge and its ability to generalize from that knowledge to new, unseen tasks.\\
\textbf{Evaluation:} This is done by presenting the model with a task description or a prompt that specifies a task, along with inputs that the model has not been explicitly prepared for. The model's output is then assessed for accuracy, relevance, or appropriateness, depending on the task.

\subsection{Application in Evaluating LLMs}
\begin{itemize}
    \item \textbf{Task Understanding}: Zero-shot learning performance evaluates an LLM's ability to understand the instructions or the task presented in natural language. This demonstrates the model's grasp of language nuances and its ability to infer the required actions without prior examples.
    \item \textbf{Generalization Capabilities}: It highlights the model's ability to apply its learned knowledge to new and diverse tasks. A high performance in zero-shot learning indicates strong generalization capabilities, a key feature for practical applications of LLMs across various domains.
    \item \textbf{Flexibility and Adaptability}: By assessing how well an LLM performs in a zero-shot setting, we can gauge its flexibility and adaptability to a broad spectrum of tasks. This is particularly valuable in real-world scenarios where it's impractical to fine-tune models for every possible task.
    \item \textbf{Semantic Understanding and Reasoning}: Zero-shot learning performance also sheds light on the model's semantic understanding and reasoning abilities. It tests whether the model can comprehend complex instructions and generate coherent, contextually appropriate responses.
\end{itemize}

\subsection{Challenges and Considerations}
\begin{itemize}
    \item \textbf{Variability in Performance}: Zero-shot learning performance can vary significantly across different tasks and domains. Some tasks may inherently align more closely with the model's training data, leading to better performance, while others may pose greater challenges.
    \item \textbf{Evaluation Criteria}: Establishing clear, objective criteria for evaluating zero-shot learning performance can be challenging, especially for subjective or open-ended tasks. It often requires carefully designed prompts and a nuanced understanding of expected outcomes.
    \item \textbf{Comparison with Few-Shot and Fine-Tuned Models}: Zero-shot learning performance is often compared against few-shot learning (where the model is given a few examples of the task) and fully fine-tuned models. This comparison helps in understanding the trade-offs between generalization and task-specific optimization.
\end{itemize}

In summary, zero-shot learning performance is a vital metric for evaluating the sophistication and applicability of LLMs. It not only underscores the models' ability to generalize across tasks without specific training but also highlights their potential for wide-ranging applications, from natural language understanding and generation to complex problem-solving across disciplines.

\section{Few-Shot Learning Performance}
Few-Shot Learning Performance is a pivotal metric for evaluating the adaptability and efficiency of Large Language Models (LLMs), such as those in the GPT series, by measuring their ability to learn and perform tasks from a minimal set of examples. This metric underscores the models' capacity to quickly generalize from limited data, a crucial attribute in scenarios with sparse training resources or when models need to adapt to new domains swiftly.

Peng (2020) introduces FewshotWOZ, a benchmark specifically designed for assessing NLG systems in task-oriented dialog contexts, showcasing the SC-GPT model's significant superiority over existing methods. Cheng (2019) explores a meta metric learning approach tailored for unbalanced classes and diverse multi-domain tasks, achieving exemplary performance in both standard and realistic few-shot learning environments. Simon (2020) discusses a dynamic classifier-based framework for few-shot learning, noting its robustness to perturbations and competitive edge in both supervised and semi-supervised few-shot classification scenarios. Additionally, Tang (2020) presents DPSN, an interpretable neural framework for few-shot time-series classification, which notably surpasses contemporary methods, especially in data-limited situations.

These contributions collectively highlight the importance and applicability of Few-Shot Learning Performance as a measure for LLMs, emphasizing the ongoing innovations and methodologies enhancing model performance under constrained learning conditions.

\subsection{Understanding Few-Shot Learning Performance}
\textbf{Concept:} Few-shot learning involves evaluating the model's ability to leverage a small number of examples to perform a task. These examples are provided to the model at inference time, typically as part of the prompt, instructing the model on the task requirements and demonstrating the desired output format or content.\\
\textbf{Evaluation:} The model's outputs are then compared against reference outputs or evaluated based on accuracy, relevance, and quality, depending on the specific task. The key is that the model uses these few examples to understand and generalize the task requirements to new, unseen instances.

\subsection{Application in Evaluating LLMs}
\begin{itemize}
    \item \textbf{Rapid Adaptation}: Few-shot learning performance showcases an LLM's ability to rapidly adapt to new tasks or domains with very little data. This is crucial for practical applications where generating or collecting large datasets for every possible task is impractical or impossible.
    \item \textbf{Data Efficiency}: This metric highlights a model's data efficiency, an important factor in scenarios where data is scarce, expensive to obtain, or when privacy concerns limit the availability of data.
    \item \textbf{Generalization from Minimal Cues}: Few-shot learning evaluates how well a model can generalize from minimal cues. It tests the model's understanding of language and task structures, requiring it to apply its pre-existing knowledge in novel ways based on a few examples.
    \item \textbf{Versatility and Flexibility}: High few-shot learning performance indicates a model's versatility and flexibility, essential traits for deploying LLMs across a wide range of tasks and domains without needing extensive task-specific data or fine-tuning.
\end{itemize}

\subsection{Challenges and Considerations}
\begin{itemize}
    \item \textbf{Consistency Across Tasks}: Few-shot learning performance can vary widely across different tasks and domains. Some tasks might naturally align with the model's pre-trained knowledge, leading to better performance, while others might be more challenging, requiring careful prompt design to achieve good results.
    \item \textbf{Quality of Examples}: The quality and representativeness of the few-shot examples significantly impact performance. Poorly chosen examples can lead to incorrect generalizations, highlighting the importance of example selection.
    \item \textbf{Comparison with Zero-Shot and Fine-Tuned Models}: Few-shot learning performance is often compared to zero-shot learning (where the model receives no task-specific examples) and fully fine-tuned models. This comparison helps in understanding the balance between adaptability and the need for task-specific optimization.
    \item \textbf{Prompt Engineering}: The effectiveness of few-shot learning can heavily depend on the skill of prompt engineering—the process of designing the prompt and examples given to the model. This skill can vary significantly among practitioners, potentially affecting the reproducibility and fairness of evaluations.
\end{itemize}

In summary, few-shot learning performance is a critical metric for evaluating the adaptability, data efficiency, and generalization capabilities of LLMs. It reflects the practical utility of these models in real-world scenarios, where the ability to perform well with limited examples is a valuable asset.

\section{Transfer Learning Evaluation}
Transfer Learning Evaluation is a key method for gauging the adaptability and efficiency of Large Language Models (LLMs) like those in the GPT series and BERT. This approach assesses an LLM's proficiency in applying pre-learned knowledge to new, related tasks without substantial additional training, highlighting the model's capability to generalize beyond its initial training parameters. Hajian (2019) underscores the significance of this evaluation, emphasizing its role in measuring a model's flexibility in applying acquired knowledge across different contexts. This method aligns with the broader educational strategies of coaching, scaffolding, and reflection in situated learning environments, further supported by Hajian (2019).

Furthermore, the evaluation extends to learning management systems (LMS) in e-learning, where factors like instruction management and screen design play critical roles (Kim, 2008). The principle of transfer of training, important for training policy and enhancing transferability, is also relevant here (Annett, 1985). Recently, the Log Expected Empirical Prediction (LEEP) metric was introduced by Nguyen (2020) as a novel measure to evaluate the transferability of learned representations, showing potential in predicting model performance and convergence speed across tasks.

This comprehensive perspective on Transfer Learning Evaluation illustrates its essential role in understanding and enhancing the utility of LLMs for a wide array of applications, from personalized learning environments to the efficient adaptation of models to new domains.

\subsection{Understanding Transfer Learning Evaluation}
\textbf{Concept:} Transfer learning involves a model applying its learned knowledge from one task (source task) to improve performance on a different but related task (target task). This process often requires minimal adjustments or fine-tuning to the model's parameters with a small dataset specific to the target task.\\
\textbf{Evaluation:} The model's performance on the target task is measured, typically using task-specific metrics such as accuracy, F1 score, BLEU score for translation tasks, or ROUGE score for summarization tasks. The improvement in performance, as compared to the model's baseline performance without transfer learning, highlights the effectiveness of the transfer learning process.

\subsection{Application in Evaluating LLMs}
\begin{itemize}
    \item \textbf{Domain Adaptation}: Transfer learning evaluation showcases an LLM's ability to adapt to specific domains or industries, such as legal, medical, or financial sectors, by applying its general language understanding to domain-specific tasks.
    \item \textbf{Efficiency in Learning}: This evaluation method highlights the model's efficiency in learning new tasks. A model that performs well in transfer learning evaluations can achieve high levels of performance on new tasks with minimal additional data or fine-tuning, indicating efficient learning and adaptation capabilities.
    \item \textbf{Model Generalization}: Transfer learning evaluation tests the generalization ability of LLMs across tasks and domains. High performance in transfer learning indicates that the model has not only memorized the training data but has also developed a broader understanding of language and tasks that can be generalized to new challenges.
    \item \textbf{Resource Optimization}: By demonstrating how well a model can adapt to new tasks with minimal intervention, transfer learning evaluation also points to the potential for resource optimization in terms of data, computational power, and time required for model training and adaptation.
\end{itemize}

\subsection{Challenges and Considerations}
\begin{itemize}
    \item \textbf{Selection of Source and Target Tasks}: The choice of source and target tasks can significantly influence the evaluation outcome. Tasks that are too similar may not adequately test the transfer capabilities, while tasks that are too dissimilar may unfairly challenge the model's ability to transfer knowledge.
    \item \textbf{Measurement of Improvement}: Quantifying the improvement and attributing it specifically to the transfer learning process can be challenging. It requires careful baseline comparisons and might need to account for variations in task difficulty and data availability.
    \item \textbf{Balancing Generalization and Specialization}: Transfer learning evaluation must balance the model's ability to generalize across tasks with its ability to specialize in specific tasks. Overemphasis on either aspect can lead to misleading conclusions about the model's overall effectiveness.
    \item \textbf{Dependency on Fine-Tuning}: The extent and method of fine-tuning for the target task can affect transfer learning performance. Over-fine-tuning may lead to overfitting on the target task, while under-fine-tuning may not fully leverage the model's transfer capabilities.
\end{itemize}

In summary, Transfer Learning Evaluation is a comprehensive approach to assess the adaptability and efficiency of LLMs in applying their pre-learned knowledge to new and related tasks. It highlights the models' potential for wide-ranging applications across various domains and tasks, demonstrating their practical utility and flexibility in real-world scenarios.

\section{Adversarial Testing}
Adversarial testing, a method used to evaluate the robustness of large language models (LLMs) against inputs designed to confuse them, has been the focus of recent research. Wang (2021) introduced Adversarial GLUE, a benchmark for assessing LLM vulnerabilities, and found that existing attack methods often produce invalid or misleading examples. Dinakarrao (2018) explored the use of adversarial training to enhance the robustness of machine learning models, achieving up to 97.65\% accuracy against attacks. Ford (2019) established a link between adversarial and corruption robustness in image classifiers, suggesting that improving one should enhance the other. Chen (2022) provided a comprehensive overview of adversarial robustness in deep learning models, covering attacks, defenses, verification, and applications. These studies collectively highlight the importance of adversarial testing in identifying and addressing vulnerabilities in LLMs and other machine learning models.

\subsection{Understanding Adversarial Testing}
\textbf{Concept:} Adversarial testing involves creating or identifying inputs that are near-misses to valid inputs but are designed to cause the model to make mistakes. These inputs can exploit the model's inherent biases, over-reliance on certain data patterns, or other weaknesses.\\
\textbf{Evaluation:} The performance of LLMs against adversarial inputs is measured, often focusing on the model's error rates, the severity of mistakes, and the model's ability to maintain coherence, relevance, and factual accuracy. The goal is to identify the model's vulnerabilities and assess its resilience.

\subsection{Application in Evaluating LLMs}
\begin{itemize}
    \item \textbf{Robustness Evaluation}: Adversarial testing is key to evaluating the robustness of LLMs, highlighting how well the model can handle unexpected or challenging inputs without compromising the quality of its outputs.
    \item \textbf{Security Assessment}: By identifying vulnerabilities, adversarial testing can inform security measures needed to protect the model from potential misuse, such as generating misleading information, bypassing content filters, or exploiting the model in harmful ways.
    \item \textbf{Bias Detection}: Adversarial inputs can reveal biases in LLMs, showing how the model might respond differently to variations in input that reflect gender, ethnicity, or other sensitive attributes, thereby guiding efforts to mitigate these biases.
    \item \textbf{Improvement of Model Generalization}: Identifying specific weaknesses through adversarial testing allows for targeted improvements to the model, enhancing its ability to generalize across a wider range of inputs and reducing overfitting to the training data.
\end{itemize}

\subsection{Challenges and Considerations}
\begin{itemize}
    \item \textbf{Generation of Adversarial Inputs}: Crafting effective adversarial inputs requires a deep understanding of the model's architecture and training data, as well as creativity to identify potential vulnerabilities. This process can be both technically challenging and time-consuming.
    \item \textbf{Measurement of Impact}: Quantifying the impact of adversarial inputs on model performance can be complex, as it may vary widely depending on the nature of the task, the model's architecture, and the specific vulnerabilities being tested.
    \item \textbf{Balance Between Robustness and Performance}: Enhancing a model's robustness to adversarial inputs can sometimes lead to trade-offs with its overall performance on standard inputs. Finding the right balance is crucial for maintaining the model's effectiveness and usability.
    \item \textbf{Ethical Considerations}: The use of adversarial testing must be guided by ethical considerations, ensuring that the insights gained are used to improve model safety and reliability, rather than for malicious purposes.
\end{itemize}

In summary, Adversarial Testing is an indispensable tool for evaluating and enhancing the robustness, security, and fairness of LLMs. By systematically challenging the models with adversarial inputs, developers can identify and address vulnerabilities, improving the models' resilience and trustworthiness in handling a wide variety of real-world applications.

\section{Fairness and Bias Evaluation}
Fairness and Bias Evaluation is crucial for assessing Large Language Models (LLMs) to ensure their outputs are equitable and free from biases that could lead to discrimination across demographics such as gender, ethnicity, age, and disability. This process not only aids in identifying biases inherent in training data or algorithms but also plays a pivotal role in mitigating potential unfair treatment. Through this evaluation, developers and researchers gain insights into the societal implications of LLMs, guiding the development of more ethical AI systems.

The significance of fairness and bias evaluation in machine learning, underscored by Mehrabi (2019) and Caton (2020), encompasses a comprehensive analysis of fairness definitions and the categorization of fairness-enhancing approaches. While Mehrabi offers a detailed taxonomy of fairness, Caton focuses on stratifying methods to promote fairness into pre-processing, in-processing, and post-processing stages. Corbett-Davies (2018) critiques these fairness definitions' statistical foundations, advocating for equitable treatment of individuals with similar risk profiles. Additionally, Pessach (2022) delves into the root causes of algorithmic bias and reviews mechanisms to improve fairness, highlighting the critical need for objective and unbiased ML algorithms. This collective body of work emphasizes the importance of rigorous fairness and bias evaluation in creating AI systems that are just and equitable.

\subsection{Understanding Fairness and Bias Evaluation}
\textbf{Concept:} This evaluation method involves analyzing the model's outputs to check for biases that may disadvantage or favor certain groups over others. It looks at how the model's predictions and responses vary across different demographic groups to identify disparities.\\
\textbf{Evaluation:} Various statistical and qualitative methods are used to measure biases in model outputs. This can include disaggregated performance metrics (such as accuracy, precision, recall) across groups, analysis of language and sentiment bias, and the use of fairness metrics like equality of opportunity, demographic parity, and others.

\subsection{Application in Evaluating LLMs}
\begin{itemize}
    \item \textbf{Identifying and Quantifying Biases}: Fairness and bias evaluation helps in identifying both explicit and implicit biases within LLM outputs. By quantifying these biases, developers can understand their extent and the specific areas where the model may need improvement.
    \item \textbf{Improving Model Generalization}: Evaluating and mitigating biases is essential for improving the generalization of LLMs. Models that perform equitably across a wide range of demographic groups are likely to be more effective and reliable in diverse real-world applications.
    \item \textbf{Enhancing Model Trustworthiness}: By addressing fairness and bias issues, developers can enhance the trustworthiness and societal acceptance of LLMs. This is particularly important for applications in sensitive areas such as healthcare, finance, and legal systems, where biased outputs can have significant consequences.
    \item \textbf{Regulatory Compliance and Ethical Standards}: Fairness and bias evaluation is critical for meeting ethical standards and regulatory requirements related to AI and machine learning. It helps ensure that LLMs adhere to principles of fairness, accountability, and transparency.
\end{itemize}

\subsection{Challenges and Considerations}
\begin{itemize}
    \item \textbf{Complexity of Bias Mitigation}: Identifying biases is only the first step; effectively mitigating them without introducing new biases or significantly impacting the model's performance is a complex challenge. It often requires iterative testing and refinement of both the model and its training data.
    \item \textbf{Multidimensional Nature of Fairness}: Fairness is a multidimensional concept that may mean different things in different contexts. Balancing various fairness criteria and understanding their implications for diverse groups can be challenging.
    \item \textbf{Data Representation and Model Transparency}: Evaluating fairness and bias often requires a deep understanding of the model's training data, algorithms, and decision-making processes. Issues of data representation and model transparency can complicate these evaluations.
    \item \textbf{Evolving Standards and Societal Norms}: Standards of what constitutes fairness and bias evolve over time and differ across cultures and communities. Continuous monitoring and updating of LLMs are necessary to align with these evolving standards.
\end{itemize}

In summary, Fairness and Bias Evaluation is critical for ensuring that LLMs are developed and deployed in a way that promotes equity and avoids harm. Through careful evaluation and ongoing efforts to mitigate identified biases, developers can contribute to the creation of more ethical and socially responsible AI systems.

\section{Robustness Evaluation}
Robustness Evaluation is vital for determining the durability and reliability of Large Language Models (LLMs) across diverse and challenging conditions, including scenarios not covered during training. This evaluation critically examines the model's capacity to sustain consistent performance amidst variations in input, adversarial attacks, and exposure to noisy data, emphasizing the importance of robustness for the safe and effective deployment of LLMs in real-world settings.

Lei (2010) and Wang (2021) underscore the significance of robustness evaluation in the LLM domain, with a focus on assessing model performance against a spectrum of challenging conditions. Wang (2021) offers an extensive survey on robustness in natural language processing (NLP), detailing various definitions, evaluation methodologies, and strategies for enhancing model robustness. Huang (2007) discusses the broader implications of robustness in product design, reinforcing the role of robust evaluation in ensuring high-quality outcomes. Additionally, Goel (2021) introduces the Robustness Gym, a unified toolkit designed for evaluating model robustness, facilitating the comparison of different evaluation approaches and contributing to the development of more resilient LLMs.

\subsection{Understanding Robustness Evaluation}
\textbf{Concept:} Robustness in the context of LLMs refers to the model's stability and reliability across diverse and unpredictable inputs. A robust model can handle variations in input data, resist manipulation through adversarial examples, and perform reliably across different domains or languages without significant degradation in performance.\\
\textbf{Evaluation:} Robustness is assessed through a series of tests designed to challenge the model in various ways. This may include:
\begin{itemize}
    \item \textbf{Input Perturbations}: Testing the model's performance on data that has been slightly altered or corrupted in ways that should not affect the interpretation for a human reader.
    \item \textbf{Adversarial Examples}: Evaluating the model against inputs specifically designed to trick or mislead it, as a way to probe for vulnerabilities.
    \item \textbf{Stress Testing}: Subjecting the model to extreme conditions, such as very long inputs, out-of-distribution data, or highly ambiguous queries, to assess its limits.
    \item \textbf{Cross-Domain Evaluation}: Testing the model's performance on data from domains or topics not covered in its training set, to assess its generalization capabilities.
\end{itemize}

\subsection{Application in Evaluating LLMs}
\begin{itemize}
    \item \textbf{Ensuring Reliability in Diverse Conditions}: Robustness evaluation helps ensure that LLMs can be deployed in a wide range of applications and environments, maintaining high performance even under conditions that differ from their training data.
    \item \textbf{Protecting Against Malicious Use}: By identifying and addressing vulnerabilities through robustness evaluation, developers can make it more difficult for malicious actors to exploit LLMs, enhancing the security of these systems.
    \item \textbf{Improving User Experience}: Ensuring robustness contributes to a better user experience by providing consistent and reliable outputs, even when users interact with the model in unexpected ways or provide noisy input data.
    \item \textbf{Facilitating Responsible Deployment}: A thorough robustness evaluation is crucial for responsibly deploying LLMs, particularly in critical applications where errors or inconsistencies could have serious consequences.
\end{itemize}

\subsection{Challenges and Considerations}
\begin{itemize}
    \item \textbf{Balancing Performance and Robustness}: Increasing a model's robustness can sometimes come at the cost of overall performance or efficiency. Finding the optimal balance is a key challenge in model development.
    \item \textbf{Comprehensive Testing}: Designing a robustness evaluation that comprehensively covers all possible challenges and conditions the model might face in real-world applications is complex and resource-intensive.
    \item \textbf{Continuous Evaluation}: The robustness of LLMs may need to be re-evaluated over time as new vulnerabilities are discovered, usage patterns evolve, or the model is applied in new contexts.
    \item \textbf{Interpretability and Diagnostics}: Understanding why a model fails under certain conditions is essential for improving robustness. However, the complexity and opacity of LLMs can make diagnosing and addressing weaknesses challenging.
\end{itemize}

In summary, Robustness Evaluation is a multifaceted approach to ensuring that LLMs are reliable, secure, and effective across a wide array of conditions and applications. By rigorously testing and improving the robustness of these models, developers can enhance their utility and safety, paving the way for their successful integration into various aspects of society and industry.

\section{LLMMaps}
LLMMaps is a pioneering visualization technique crafted for the nuanced evaluation of Large Language Models (LLMs) within various NLP subfields. It seeks to offer an all-encompassing assessment of an LLM's performance, highlighting both its strengths and areas requiring improvement, particularly focusing on reducing hallucinations—where models erroneously present incorrect information as accurate. Puchert (2023) underscores the value of LLMMaps in detecting performance discrepancies and susceptibility to hallucinations in LLMs. Complementing this, CRITIC, introduced by Gou (2023), enables LLMs to self-correct via interactions with external tools. Furthermore, Peng (2023) proposes enhancing LLMs with external knowledge and automated feedback to further curb hallucinations. Collectively, these strategies aim to bolster the precision and dependability of LLMs, marking significant progress in NLP technology.

\subsection{Understanding LLMMaps}
\textbf{Concept:} LLMMaps organizes and visualizes the performance of LLMs across a spectrum of NLP tasks and domains in a structured manner. This stratification allows researchers and developers to pinpoint specific areas of excellence and those in need of refinement.\\
\textbf{Visualization:} The technique could involve graphical representations, such as heatmaps or multidimensional plots, where each axis or dimension corresponds to different evaluation criteria or NLP subfields. Performance metrics, such as accuracy, fairness, robustness, or the propensity for hallucinations, can be represented in this multidimensional space.\\
\textbf{Hallucination Focus:} A significant aspect of LLMMaps is its emphasis on identifying and reducing hallucinations. By visualizing areas where hallucinations are more prevalent, developers can target improvements more effectively.

\subsection{Application in Evaluating LLMs}
\begin{itemize}
    \item \textbf{Comprehensive Performance Overview:} LLMMaps can provide a holistic view of an LLM's performance, highlighting how well it performs across a variety of tasks, such as translation, summarization, question-answering, and more. This overview helps in understanding the model's general capabilities and limitations.
    \item \textbf{Targeted Improvements:} By visually identifying areas requiring improvement, such as those prone to hallucinations or biases, LLMMaps enables developers to focus their efforts more effectively on enhancing model quality and reliability.
    \item \textbf{Benchmarking and Comparison:} LLMMaps can be used as a benchmarking tool, allowing for the comparison of different models or versions of a model over time. This can track progress and inform the development of more advanced, less error-prone models.
    \item \textbf{Facilitating Research and Collaboration:} The visual and stratified nature of LLMMaps makes it an excellent tool for facilitating discussions and collaborations within the research community, helping to align efforts towards addressing common challenges.
\end{itemize}

\subsection{Challenges and Considerations}
\begin{itemize}
    \item \textbf{Data and Metric Selection:} The effectiveness of LLMMaps depends on the selection of relevant data and metrics for evaluation. Ensuring these are comprehensive and accurately reflect model performance is crucial.
    \item \textbf{Complexity in Interpretation:} While LLMMaps can provide a detailed overview of model performance, interpreting these visualizations, especially in highly multidimensional spaces, can be complex and require expertise in data analysis and visualization techniques.
    \item \textbf{Updating and Maintenance:} As the field of NLP evolves, maintaining LLMMaps to reflect new subfields, evaluation metrics, and challenges will be necessary to keep them relevant and useful.
    \item \textbf{Subjectivity and Bias:} The design and interpretation of LLMMaps might introduce subjectivity, especially in how performance areas are defined and prioritized. Ensuring objectivity and inclusiveness in these evaluations is important to avoid reinforcing existing biases.
\end{itemize}

In summary, LLMMaps represent a novel and potentially powerful approach to evaluating LLMs, offering detailed insights into their performance across various dimensions. By highlighting specific areas for improvement, especially in reducing hallucinations, LLMMaps can guide the development of more accurate, reliable, and fair LLMs.

\section{Benchmarking and Leaderboards}
Benchmarking and Leaderboards serve as crucial instruments for systematically assessing the performance of Large Language Models (LLMs), particularly in their ability to address queries from extensive Q\&A datasets. Hockney (1993) emphasizes the importance of selecting appropriate performance metrics, cautioning against the reliance on speedup and MMop/s measures due to their potential limitations in capturing the nuanced capabilities of LLMs. In response to the demand for more rigorous benchmarks, Arora (2023) introduced JEEBench, a collection of intricate problems demanding extended reasoning and specialized knowledge. This benchmark has highlighted the advancements in newer LLMs, while also pointing out areas needing further development. Additionally, Vestal (1990) suggested a method for benchmarking language features through multiple sampling loops and linear regression, a technique that could offer detailed performance insights for various LLM parameters. Collectively, these approaches underscore the role of Benchmarking and Leaderboards in evaluating LLMs, pushing the envelope for accuracy and proficiency in complex language understanding tasks.

\subsection{Understanding Benchmarking and Leaderboards}
\textbf{Benchmarking:} This involves evaluating LLMs against a standardized set of tasks or datasets to measure their performance. In the context of Q\&A, benchmark datasets consist of a large number of questions paired with correct answers, covering various topics and difficulty levels. The model's responses are compared to the correct answers to assess accuracy, comprehension, and relevance.\\
\textbf{Leaderboards:} Leaderboards rank LLMs based on their performance on benchmark tasks. They provide a comparative view of different models, highlighting which models perform best on specific tasks or datasets. Leaderboards are often hosted by academic conferences, research institutions, or industry groups, and they are updated regularly as new models are developed and evaluated.

\subsection{Application in Evaluating LLMs}
\begin{itemize}
    \item \textbf{Performance Assessment:} Benchmarking and leaderboards offer a clear, quantitative measure of an LLM's ability to understand and process natural language queries, providing insights into its comprehension, reasoning, and language generation capabilities.
    \item \textbf{Model Comparison:} By placing models in a competitive context, leaderboards help identify the most advanced LLMs in terms of Q\&A accuracy and other metrics, fostering a healthy competition among researchers and developers to improve their models.
    \item \textbf{Progress Tracking:} Benchmarking allows for the tracking of progress in the field of NLP and LLM development over time. It shows how models evolve and improve, indicating advancements in technology and methodologies.
    \item \textbf{Identifying Strengths and Weaknesses:} Through detailed analysis of benchmarking results, developers can identify specific areas where their models excel or fall short, informing targeted improvements and research directions.
\end{itemize}

\subsection{Challenges and Considerations}
\begin{itemize}
    \item \textbf{Diversity and Representativeness:} Ensuring that benchmark datasets are diverse and representative of real-world questions is crucial for meaningful evaluation. Biases or limitations in the datasets can lead to skewed assessments of model capabilities.
    \item \textbf{Beyond Accuracy:} While accuracy is a critical metric, it does not capture all aspects of an LLM's performance. Other factors like response time, resource efficiency, and the ability to generate nuanced, context-aware responses are also important.
    \item \textbf{Dynamic Nature of Leaderboards:} As new models are constantly being developed, leaderboards are always changing. Staying at the top of a leaderboard can be fleeting, emphasizing the need for continuous improvement and adaptation.
    \item \textbf{Overemphasis on Competition:} While competition can drive innovation, excessive focus on leaderboard rankings may lead to over-optimization for specific benchmarks at the expense of generalizability and ethical considerations.
\end{itemize}

In summary, Benchmarking and Leaderboards are invaluable tools for evaluating LLMs, especially in the domain of question answering. They provide a structured and competitive environment for assessing model performance, driving advancements in the field. However, it's important to consider these tools as part of a broader evaluation strategy that also includes qualitative assessments, ethical considerations, and real-world applicability to fully understand and improve the capabilities of LLMs.

\section{Stratified Analysis}
Stratified Analysis is a versatile evaluation method that dissects Large Language Models' (LLMs) performance into distinct layers or strata, each representing various domains, topics, or task types. This granular approach allows for a detailed understanding of LLMs' strengths and weaknesses across different knowledge subfields. The concept of stratified analysis, while diverse in application, shares a common goal of providing in-depth insights within specific contexts. Moutinho (1994) introduced Stratlogic, a strategic marketing tool that analyzes competitive positioning through a data-driven lens. Kumar (1997) assessed data formats in layered manufacturing, detailing their advantages and limitations. Rahwan (2007) developed STRATUM, a strategy for designing heuristic negotiation tactics in automated negotiations, underscoring the need to account for agent capabilities. Jongman (2005) applied statistical environmental stratification across Europe, aiming to streamline environmental patterns for improved biodiversity assessment and monitoring. Together, these applications underscore the broad utility and adaptability of stratified analysis in enhancing domain-specific understanding and strategy development.

\subsection{Understanding Stratified Analysis}
\textbf{Concept:} Stratified analysis breaks down the evaluation of LLMs into smaller, more manageable segments based on predefined criteria such as content domains (e.g., science, literature, technology), task types (e.g., question answering, summarization, translation), or complexity levels. This allows for a detailed assessment of the model's performance in each area.\\
\textbf{Application:} The performance of an LLM is assessed within each stratum using relevant metrics, such as accuracy, precision, recall, or domain-specific evaluation standards. This detailed assessment helps in understanding how well the model handles different types of information and tasks.

\subsection{Application in Evaluating LLMs}
\begin{itemize}
    \item \textbf{Identifying Domain-Specific Performance:} Stratified analysis enables the identification of which domains or topics an LLM excels in and which it struggles with. For instance, a model might perform exceptionally well in technical domains but poorly in creative writing or ethical reasoning.
    \item \textbf{Guiding Model Improvements:} By pinpointing specific areas of weakness, this analysis directs researchers and developers towards targeted improvements, whether by adjusting training data, refining model architectures, or incorporating specialized knowledge sources.
    \item \textbf{Enhancing Generalization and Specialization:} Understanding a model's performance across various strata can inform strategies for enhancing its generalization capabilities while also developing specialized models tailored for specific domains or tasks.
    \item \textbf{Benchmarking and Comparative Analysis:} Stratified analysis facilitates more nuanced benchmarking and comparison between models, allowing for a deeper understanding of each model's unique strengths and limitations in a variety of contexts.
\end{itemize}

\subsection{Challenges and Considerations}
\begin{itemize}
    \item \textbf{Selection of Strata:} Determining the appropriate strata for analysis can be challenging. The criteria for stratification need to be carefully chosen to ensure that the analysis is meaningful and covers the breadth of knowledge and tasks relevant to LLMs.
    \item \textbf{Comprehensive Evaluation:} Conducting a thorough stratified analysis requires significant resources, including diverse datasets and domain-specific evaluation metrics. Ensuring comprehensiveness while managing these resources is a key challenge.
    \item \textbf{Balancing Depth and Breadth:} While stratified analysis offers depth in specific areas, it's essential to balance this with a broad overview to avoid missing the bigger picture of the model's capabilities.
    \item \textbf{Evolving Knowledge Fields:} As knowledge and technology evolve, the strata used for analysis may need to be updated or expanded, requiring ongoing adjustments to evaluation frameworks.
\end{itemize}

In summary, Stratified Analysis offers a detailed and nuanced approach to evaluating LLMs, shedding light on their varied capabilities across different domains and tasks. This method provides valuable insights that can guide the development of more capable, versatile, and targeted LLMs, ultimately advancing the field of natural language processing and artificial intelligence.

\section{Visualization of Bloom’s Taxonomy}
A range of studies have explored the application of Bloom's Taxonomy in different contexts. Granello (2001) and Köksal (2018) both emphasize the importance of this framework in education, with Granello focusing on its use in graduate-level writing and Köksal in language assessment. Kelly (2006) and Yusof (2010) delve into the practical aspects of applying Bloom's Taxonomy, with Kelly proposing a context-aware analysis scheme and Yusof developing a classification model for question items in examinations. These studies collectively highlight the versatility and potential of Bloom's Taxonomy as a tool for enhancing cognitive complexity and evaluating performance.

\subsection{Understanding the Visualization of Bloom's Taxonomy}
\textbf{Concept:} This approach visualizes the model's performance in a pyramidal (or hierarchical) fashion, reflecting the structure of Bloom's Taxonomy itself. Each level of the pyramid represents a level of cognitive skill, with the base indicating tasks that require basic memory (Remember) and the apex representing tasks that require creative abilities (Create).\\
\textbf{Application:} The accuracy or performance metric for the LLM is calculated for tasks aligned with each of Bloom's levels. These metrics are then plotted on the pyramid, allowing for a clear visual representation of where the model excels or struggles.

\subsection{Application in Evaluating LLMs}
\begin{itemize}
    \item \textbf{Assessing Cognitive Capabilities:} This visualization helps in understanding the range and depth of cognitive tasks an LLM can handle. For instance, a model may perform well in tasks that require understanding and applying knowledge but struggle with tasks requiring evaluation and creation.
    \item \textbf{Guiding Model Development:} By identifying specific cognitive levels where the LLM's performance is lacking, developers can focus their efforts on improving these areas, whether through training on more diverse datasets, incorporating advanced algorithms, or integrating additional knowledge sources.
    \item \textbf{Educational Applications:} For LLMs intended for educational purposes, visualization of Bloom's Taxonomy can be particularly useful in aligning the model's capabilities with educational goals and standards, ensuring it supports learning across all cognitive levels.
    \item \textbf{Benchmarking Complexity Handling:} This method provides a standardized way to benchmark and compare the sophistication of different LLMs in handling tasks of varying cognitive complexity, offering a comprehensive view of their intellectual capabilities.
\end{itemize}

\subsection{Challenges and Considerations}
\begin{itemize}
    \item \textbf{Task Alignment:} Aligning tasks with the appropriate level of Bloom's Taxonomy can be subjective and requires a deep understanding of both the taxonomy and the tasks being evaluated. Misalignment could lead to inaccurate assessments of model capabilities.
    \item \textbf{Complexity of Evaluation:} Tasks at higher cognitive levels (e.g., Evaluate, Create) are inherently more complex and subjective, making them challenging to evaluate accurately. Developing reliable metrics for these tasks is crucial for meaningful visualization.
    \item \textbf{Interpretation of Results:} While the visualization provides a clear overview of performance across cognitive levels, interpreting these results and translating them into actionable insights requires careful consideration of the model's intended applications and limitations.
    \item \textbf{Dynamic Nature of LLM Capabilities:} As LLMs evolve and improve, their capabilities at different levels of Bloom's Taxonomy may change. Ongoing evaluation and updating of the visualization are necessary to maintain an accurate representation of their performance.
\end{itemize}

In summary, Visualization of Bloom's Taxonomy offers a unique and insightful method for evaluating LLMs, highlighting their capabilities and limitations across a spectrum of cognitive tasks. This approach not only aids in the targeted development of LLMs but also in their application in educational and complex problem-solving contexts, pushing the boundaries of what these models can achieve.

\section{Hallucination Score}
The phenomenon of hallucinations in Large Language Models (LLMs)—where models generate unfounded or entirely fictional responses—has emerged as a significant concern, compromising the reliability and trustworthiness of AI systems. Highlighted by researchers like Ye (2023) and Lee (2018), these inaccuracies can severely impact LLM applications, from educational tools to critical news dissemination. In response, Zhou (2020) introduced a novel technique for identifying hallucinated content in neural sequence generation, marking a pivotal step towards enhancing sentence-level hallucination detection and significantly improving the reliability of LLM outputs.

Within this context, the Hallucination Score, a metric developed as part of the LLMMaps framework, plays a crucial role by measuring the frequency and severity of hallucinations in LLM outputs. This metric enables a systematic assessment of how often and to what extent LLMs produce unsupported or incorrect responses, guiding efforts to mitigate such issues and bolster the models' applicability in sensitive and critical domains.

\subsection{Understanding the Hallucination Score}
\textbf{Concept:} The Hallucination Score measures the extent to which an LLM produces hallucinated content. It is quantified based on the analysis of the model's outputs against verified information or established facts, considering both the frequency of hallucinations and their potential impact.\\
\textbf{Application:} To calculate this score, responses from the LLM are evaluated against a benchmark set of questions or prompts that have known, factual answers. The score might be derived from the proportion of responses that contain hallucinations, weighted by the severity or potential harm of the inaccuracies.

\subsection{Application in Evaluating LLMs}
\begin{itemize}
    \item \textbf{Identifying Reliability Issues:} By quantifying hallucinations, the score helps in identifying how often and under what conditions an LLM might produce unreliable outputs. This is crucial for assessing the model's suitability for various applications.
    \item \textbf{Guiding Model Improvements:} A high Hallucination Score signals a need for model refinement, possibly through better training data curation, improved model architecture, or enhanced post-processing checks to minimize inaccuracies.
    \item \textbf{Benchmarking and Comparison:} The Hallucination Score provides a standardized metric for comparing different models or versions of a model over time, offering insights into progress in reducing hallucinations and improving output accuracy.
    \item \textbf{Enhancing User Trust:} By actively monitoring and working to reduce the Hallucination Score, developers can enhance user trust in LLM applications, ensuring that the information provided is accurate and reliable.
\end{itemize}

\subsection{Challenges and Considerations}
\begin{itemize}
    \item \textbf{Subjectivity in Evaluation:} Determining what constitutes a hallucination can be subjective, especially in areas where information is ambiguous or rapidly evolving. Developing clear criteria for identifying and categorizing hallucinations is essential.
    \item \textbf{Complexity of Measurement:} Accurately measuring the Hallucination Score requires comprehensive evaluation across a wide range of topics and contexts, necessitating significant resources and expert knowledge.
    \item \textbf{Balancing Creativity and Accuracy:} In some applications, such as creative writing or idea generation, a certain level of "hallucination" might be desirable. Balancing the need for creativity with the need for factual accuracy is a nuanced challenge.
    \item \textbf{Dynamic Nature of Knowledge:} As new information becomes available and the world changes, responses that were once considered accurate may become outdated or incorrect. Continuous updating and re-evaluation are necessary to maintain the validity of the Hallucination Score.
\end{itemize}

In summary, the Hallucination Score within the LLMMaps framework provides a valuable metric for evaluating the accuracy and reliability of LLM outputs. By quantifying the extent of hallucinated content, it offers a clear indicator of a model's current performance and areas for improvement, contributing to the development of more trustworthy and effective LLMs.

\section{Knowledge Stratification Strategy}
The Knowledge Stratification Strategy is a systematic evaluative method aimed at enhancing the analysis of Large Language Models (LLMs) through the organization of Q\&A datasets into a hierarchical knowledge structure. This approach categorizes questions and answers by their knowledge complexity and specificity, arranging them from broad, general knowledge at the top to highly specialized knowledge at the bottom. Such stratification facilitates a detailed analysis of an LLM's performance across various levels of knowledge depth and domain specificity, providing insights into the model's proficiency in different areas.

Drawing parallels with established methodologies in other fields, this strategy echoes the Knowledge Partitioning approach in Product Lifecycle Management (PLM) described by Therani (2005), which organizes organizational knowledge into distinct categories. It also aligns with the method used for the statistical environmental stratification of Europe by Jongman (2005), aimed at delineating environmental gradients for better assessment. In the context of the service sector, specifically IT services, Gulati (2014) highlights its importance for effective knowledge retention and management. Furthermore, Folkens (2004) discusses its application in evaluating Knowledge Management Systems (KMS) within organizations, underscoring the strategy's versatility and utility across diverse domains.

\subsection{Understanding Knowledge Stratification Strategy}
\textbf{Concept:} This strategy creates a layered framework within Q\&A datasets, where each layer represents a different level of knowledge complexity and domain specialization. The top layers might include questions that require common knowledge and understanding, while lower layers would contain questions necessitating deep, specific expertise.\\
\textbf{Application:} In evaluating LLMs, questions from different strata of the hierarchy are posed to the model. The model's performance on these questions is then analyzed to determine how well it handles various types of knowledge, from the most general to the most specialized.

\subsection{Application in Evaluating LLMs}
\begin{itemize}
    \item \textbf{Comprehensive Performance Insight:} The Knowledge Stratification Strategy offers a comprehensive view of an LLM's performance spectrum, showcasing its proficiency in handling both general and specialized queries. This insight is crucial for applications requiring a broad range of knowledge.
    \item \textbf{Identifying Areas for Improvement:} By pinpointing the levels of knowledge where the LLM's performance dips, this strategy guides targeted improvements, whether in training data augmentation, model fine-tuning, or incorporating external knowledge bases.
    \item \textbf{Enhancing Domain-Specific Applications:} For LLMs intended for domain-specific applications, this approach helps in assessing and enhancing their expertise in the relevant knowledge areas, ensuring they meet the required standards of accuracy and reliability.
    \item \textbf{Benchmarking and Comparison:} Knowledge Stratification enables a more detailed benchmarking process, allowing for the comparison of LLMs not just on overall accuracy but on their ability to navigate and respond across a spectrum of knowledge depths.
\end{itemize}

\subsection{Challenges and Considerations}
\begin{itemize}
    \item \textbf{Hierarchy Design:} Designing an effective knowledge hierarchy requires a deep understanding of the subject matter and the relevant domains, posing a challenge in ensuring the stratification is meaningful and accurately reflects varying knowledge depths.
    \item \textbf{Evaluation Consistency:} Ensuring consistent evaluation across different knowledge strata can be challenging, especially when dealing with specialized knowledge areas where expert validation may be necessary.
    \item \textbf{Adaptation to Evolving Knowledge:} The knowledge landscape is constantly evolving, particularly in specialized fields. The stratification strategy must be adaptable to incorporate new developments and discoveries, requiring ongoing updates to the hierarchy.
    \item \textbf{Balance Between Generalization and Specialization:} While stratification helps in assessing specialized knowledge, it's also important to maintain a balance, ensuring the LLM remains versatile and effective across a wide range of topics and not just narrowly focused areas.
\end{itemize}

In summary, the Knowledge Stratification Strategy offers a structured and in-depth approach to evaluating LLMs, allowing for a detailed assessment of their capabilities across a hierarchical spectrum of knowledge. By leveraging this strategy, developers and researchers can gain valuable insights into the strengths and weaknesses of LLMs, guiding the development of models that are both versatile and deeply knowledgeable in specific domains.

\section{Utilization of Machine Learning Models for Hierarchy Generation}
Utilizing Machine Learning Models for Hierarchy Generation offers a sophisticated method for structuring and analyzing Q\&A datasets to evaluate Large Language Models (LLMs). This technique employs LLMs and other machine learning models to autonomously classify and arrange questions into a coherent hierarchy of topics and subfields, ensuring each question is accurately categorized by its content and the overarching themes of the dataset. This process enhances the systematic and detailed evaluation of LLMs.

Research in this domain includes Gaussier (2002), who introduced a hierarchical generative model aimed at clustering and document categorization, aligning with the goals of hierarchy generation. Xu (2018) expanded on this by integrating prior knowledge into building topic hierarchies, offering a more refined approach. Dorr (1998) contributed a thematic hierarchy designed for efficient generation from lexical-conceptual structures, aiding in the organization of information. Ruiz (2004) explored text categorization using a hierarchical array of neural networks, showcasing the approach's utility in bolstering categorization performance. Together, these studies underscore the effectiveness and versatility of machine learning models in creating structured hierarchies for enhancing LLM evaluation and beyond.

\subsection{Understanding Utilization of Machine Learning Models for Hierarchy Generation}
\textbf{Concept:} This approach uses machine learning algorithms, including LLMs themselves, to analyze the content and context of questions in a dataset. The model identifies key themes, topics, and the complexity level of each question, using this information to generate a hierarchical structure that organizes questions

\section{Shapley Values for LLMs}
Shapley Values, derived from cooperative game theory, present a refined method for assessing the contribution of individual input features, like words or tokens, to the outputs of Large Language Models (LLMs). This technique assigns a quantifiable value to each feature based on its impact on the model's predictions, enabling a detailed examination of feature importance. By applying Shapley values to LLMs, we can achieve a deeper understanding of how each element of input data influences the model's outputs, providing a fair and robust measure of the significance of different aspects of the input.

The utility of Shapley values extends beyond LLMs, finding applications in various machine learning facets, including feature selection, model explainability, and data valuation, as explored by Rozemberczki (2022). This approach not only enhances our grasp of feature importance in LLMs but also contributes to equitable solutions in other sectors, such as fair transmission cost allocation in competitive power markets (Tan, 2002), and broadens its applicability to scenarios involving both transferable and non-transferable utility (Aumann, 1994). Through these applications, Shapley values offer a comprehensive framework for dissecting and understanding the intricate dynamics at play in LLMs and other complex systems.

\subsection{Understanding Shapley Values in the Context of LLMs}
\textbf{Equitable Distribution of Contribution:} Shapley values calculate the average marginal contribution of each feature across all possible combinations of features. This ensures that the contribution of each input feature is fairly assessed, taking into account the presence or absence of other features.\\
\textbf{Quantifying Feature Importance:} By applying Shapley values to LLMs, researchers can quantitatively determine how much each word or token in the input text contributes to the model's output. This is particularly valuable in tasks where understanding the influence of specific linguistic elements is crucial, such as sentiment analysis, text classification, or machine translation.\\
\textbf{Insights into Model Behavior:} Shapley values can reveal insights into the model's behavior, such as dependencies between features or the significance of specific words in context. This can help identify whether the model is focusing on relevant information or being swayed by irrelevant details.

\subsection{Application in LLM Evaluation}
\begin{itemize}
    \item \textbf{Model Interpretability:} Enhancing the interpretability of LLMs is one of the key applications of Shapley values. By providing a clear and fair attribution of output contributions to input features, they help demystify the model's decision-making process, making it more accessible and understandable to humans.
    \item \textbf{Bias Detection and Mitigation:} Shapley values can help identify biases in model predictions by highlighting input features that disproportionately affect the output. This can guide efforts to mitigate these biases, either by adjusting the training data or modifying the model architecture.
    \item \textbf{Improving Model Robustness:} Understanding feature contributions can inform the development of more robust LLMs. If certain innocuous features are found to have an outsized impact on predictions, this may indicate vulnerabilities to adversarial attacks or overfitting, which can then be addressed.
\end{itemize}

\subsection{Techniques and Considerations}
\begin{itemize}
    \item \textbf{Computational Complexity:} One of the challenges of applying Shapley values to LLMs is their computational intensity. Calculating the contribution of each feature requires evaluating the model's output across all possible subsets of features, which can be prohibitively expensive for large models and inputs.
    \item \textbf{Approximation Methods:} To mitigate computational challenges, various approximation algorithms have been developed. These methods aim to provide accurate estimations of Shapley values without exhaustive computation, making the approach more feasible for practical applications.
    \item \textbf{Integration with Other Interpretability Tools:} Shapley values can be used in conjunction with other interpretability tools, such as attention visualization or sensitivity analysis, to provide a more comprehensive understanding of model behavior. Combining methods can offer both detailed feature-level insights and broader overviews of model dynamics.
\end{itemize}

Shapley values represent a powerful tool for dissecting and understanding the contributions of individual features in LLM outputs. Despite their computational demands, the depth and fairness of the insights they provide make them an invaluable asset for enhancing the transparency, fairness, and interpretability of LLMs. As LLMs continue to evolve and their applications become increasingly widespread, techniques like Shapley values will play a crucial role in ensuring these models are both understandable and accountable.

\section{Attention Visualization}
Attention Visualization serves as a key technique for interpreting Large Language Models (LLMs), particularly those built on the Transformer architecture, by revealing how these models allocate importance to various parts of the input data through attention mechanisms. This visualization helps elucidate the model's focus areas within the input text, offering a window into its information processing strategies and decision-making patterns.

The concept of visual attention, as initially proposed by Tsotsos (1995) through a selective tuning model, underscores the efficiency of focusing on specific parts of the visual field. This foundational idea parallels the selective focus enabled by attention mechanisms in LLMs, especially Transformers, which adjust their focus dynamically across the input data to enhance processing efficiency. Yang (2021) advanced this concept within vision transformer models, addressing local region prediction inconsistencies by refining self-attention mechanisms. Ilinykh (2022) delved into multi-modal transformers, analyzing how cross-attention layers capture syntactic, semantic, and visual grounding information. Furthermore, Gao (2022) introduced an Attention in Attention (AiA) module aimed at refining attention correlations, thereby boosting visual tracking performance.

Collectively, these contributions from Tsotsos (1995), Yang (2021), Ilinykh (2022), and Gao (2022) enrich our understanding of attention mechanisms' role in both human cognition and artificial intelligence, highlighting the evolution and optimization of these systems in LLMs. By visualizing attention weights, researchers can dissect and improve how LLMs prioritize information, enhancing model interpretability and effectiveness.

\subsection{Understanding Attention Visualization in LLMs}
\textbf{Mechanics of Attention:} In the context of LLMs, the attention mechanism allows the model to allocate varying degrees of "focus" or "importance" to different input elements when performing a task. This mechanism is key to the model's ability to handle long-range dependencies and contextual nuances in text.\\
\textbf{Visualization Techniques:} Attention visualization typically involves creating heatmaps or other graphical representations that show the attention scores between different parts of the input text or between the input and output tokens. High attention scores are often highlighted in warmer colors (e.g., reds), indicating areas of the text that the model pays more attention to during its processing.

\subsection{Application in LLM Evaluation}
\begin{itemize}
    \item \textbf{Insights into Model Decision-making:} Visualization of attention weights provides a direct window into the decision-making process of LLMs. It can reveal how the model prioritizes certain words or phrases over others, offering clues about its understanding of language and context.
    \item \textbf{Understanding Contextual Processing:} Attention patterns can demonstrate how the model handles context, showing whether and how it integrates contextual information from different parts of the text to generate coherent and contextually appropriate responses.
    \item \textbf{Improving Model Interpretability:} By making the model's focus areas explicit, attention visualization enhances the interpretability of LLMs. This can be particularly useful for developers and researchers looking to debug or improve model performance, as well as for end-users seeking explanations for model outputs.
    \item \textbf{Identifying Biases and Artifacts:} Analyzing attention distributions can also help identify potential biases or training artifacts that the model may have learned. For instance, if the model consistently pays undue attention to specific tokens or phrases that are not relevant to the task, it might indicate a bias introduced during training.
\end{itemize}

\subsection{Techniques and Considerations}
\begin{itemize}
    \item \textbf{Layer-wise and Head-wise Visualization:} Modern Transformer-based LLMs contain multiple layers and heads within their attention mechanisms. Visualizing attention across different layers and heads can provide a more granular understanding of how information is processed and integrated at various stages of the model.
    \item \textbf{Quantitative Analysis:} Beyond visual inspection, quantitative analysis of attention weights can offer additional insights. For instance, aggregating attention scores across a dataset can highlight general patterns or biases in how the model processes different types of input.
    \item \textbf{Interpretation Challenges:} While attention visualization is a powerful tool, interpreting these visualizations can be challenging. High attention does not always equate to causal importance, and the relationship between attention patterns and model outputs can be complex.
    \item \textbf{Complementary Tools:} To gain a comprehensive understanding of LLM behavior, attention visualization is often used in conjunction with other interpretability and evaluation techniques, such as feature importance methods, Shapley values, and sensitivity analysis.
\end{itemize}

Attention Visualization stands out as a valuable technique for demystifying the complex processing mechanisms of LLMs, offering both researchers and practitioners a way to visually interrogate and understand the model's focus and decision-making processes. Through careful analysis and interpretation of attention patterns, one can derive actionable insights to enhance model performance, fairness, and user trust.

\section{Counterfactual Explanations for LLMs}
Counterfactual Explanations are a pivotal interpretability technique for Large Language Models (LLMs), focusing on how slight modifications to input data affect the model's outputs. This method, which entails exploring "what if" scenarios, is instrumental in unveiling the conditions that prompt changes in the model's decisions or predictions, thereby illuminating its underlying reasoning and causal mechanisms.

Galles (1998) and Roese (1997) highlight the importance of counterfactual explanations in understanding an LLM's decision-making process by observing the outcomes of minor changes to inputs. Höfler (2005) emphasizes the significance of a causal interpretation of counterfactuals, especially in recursive models, for gaining insights into the model's logic. Meanwhile, Briggs (2012) discusses the ongoing debate around the causal modeling semantics for counterfactuals versus the similarity-based semantics proposed by Lewis, indicating the complexity and depth of understanding required to effectively apply counterfactual explanations to LLMs.

Through these references, the value of counterfactual explanations in dissecting and comprehending the decision-making processes of LLMs is underscored, showcasing their role in enhancing model transparency and interpretability.

\subsection{Application in LLM Evaluation}
\begin{itemize}
    \item \textbf{Unveiling Model Sensitivity:} Counterfactual explanations reveal the sensitivity of LLMs to different parts of the input text. By changing certain words or phrases and observing the impact on the output, evaluators can identify which aspects of the input are most influential in the model's decisions or predictions.
    \item \textbf{Understanding Decision Boundaries:} This technique helps delineate the conditions and boundaries within which the model's output changes. It can highlight the thresholds of change necessary for the model to alter its response, offering insights into the model's internal logic and how it discriminates between different inputs.
    \item \textbf{Identifying Bias and Ethical Concerns:} By creating counterfactuals that alter demographic or contextually sensitive aspects of the input, researchers can uncover biases in the model's outputs. This is instrumental in evaluating the fairness of LLMs and identifying potential ethical issues arising from biased or stereotypical responses.
    \item \textbf{Enhancing Model Robustness:} Counterfactual explanations can also be used to test the robustness of LLMs against adversarial inputs or to ensure consistency in the model's reasoning across similar yet slightly varied inputs. This can guide efforts to improve the model's resilience to input variations and adversarial attacks.
\end{itemize}

\subsection{Techniques and Considerations}
\begin{itemize}
    \item \textbf{Minimal and Relevant Changes:} Effective counterfactual explanations typically involve minimal but meaningful changes to the input, ensuring that the observed differences in output are attributable to specific modifications. This requires a careful selection of input alterations that are relevant to the model's task and the aspect of performance being evaluated.
    \item \textbf{Systematic Generation of Counterfactuals:} Generating counterfactuals can be approached systematically by using algorithms that identify or create variations of the input data, which are likely to produce significant changes in the output. Techniques such as gradient-based optimization or genetic algorithms can automate the generation of impactful counterfactuals.
    \item \textbf{Qualitative and Quantitative Analysis:} The evaluation of counterfactual explanations involves both qualitative analysis (e.g., assessing changes in the sentiment or theme of the output) and quantitative measures (e.g., differences in output probabilities or confidence scores). Combining these approaches provides a richer understanding of the model's behavior.
    \item \textbf{Contextual and Cultural Considerations:} When creating counterfactuals, it's crucial to consider the context and cultural implications of the input changes. Misinterpretations or oversights in these areas can lead to misleading conclusions about the model's performance and decision-making process.
\end{itemize}

\subsection{Challenges}
\begin{itemize}
    \item \textbf{Interpretation Complexity:} Interpreting the results of counterfactual explanations can be challenging, especially when dealing with complex or ambiguous inputs and outputs. It requires a nuanced understanding of both the domain and the model's capabilities.
    \item \textbf{Scalability:} Manually creating and analyzing counterfactuals for a large number of inputs can be time-consuming and may not be scalable for extensive evaluations. Automation techniques can help, but they require careful design to ensure the relevance and effectiveness of the generated counterfactuals.
\end{itemize}

Counterfactual Explanations offer a powerful means to probe the inner workings of LLMs, providing valuable insights into their sensitivity, decision-making boundaries, and potential biases. By methodically exploring how changes in the input influence the output, evaluators can enhance their understanding of LLM behavior, leading to more transparent, fair, and robust language models.

\section{Language-Based Explanations for LLMs}
Language-Based Explanations (LBEs) are a vital method for making Large Language Models (LLMs) more understandable by translating their decision-making processes into natural language, accessible to humans. This approach, which can involve either the LLM itself or a dedicated model, breaks down the complex operations of machine learning into explanations that are easy for non-experts to grasp, enhancing transparency and trust in AI applications.

Celikyilmaz (2012) highlights the significance of LBEs in improving LLM interpretability. Further, the Language Interpretability Tool (LIT) introduced by Tenney (2020) offers a practical solution for visualizing and dissecting the workings of NLP models, including LLMs. Additionally, knowledge representation systems like LLILOG (Pletat, 1992) facilitate the conversion of natural language texts into formats that machines can process, underpinning the generation of language-based explanations. Wen (2015) demonstrates the impact of semantically conditioned LSTM-based natural language generation on enhancing spoken dialogue systems, illustrating a key area where LLMs benefit from improved performance through interpretability.

Together, these references emphasize the crucial role of LBEs in bridging the gap between the advanced computational abilities of LLMs and the need for their outputs to be understandable and actionable for human users, thereby making AI technologies more accessible and interpretable.

\subsection{Application in LLM Evaluation}
\begin{itemize}
    \item \textbf{Enhancing Interpretability and Transparency:} By generating explanations in natural language, LLMs become more transparent, allowing users and developers to understand the rationale behind specific outputs. This transparency is crucial for building trust and facilitating the broader adoption of LLM technologies in sensitive or critical applications.
    \item \textbf{Facilitating Debugging and Model Improvement:} Language-based explanations can highlight unexpected or erroneous reasoning patterns, serving as a valuable tool for debugging and refining LLMs. Understanding why a model produces a particular output enables targeted interventions to correct biases, improve accuracy, and enhance overall performance.
    \item \textbf{Supporting Ethical AI Practices:} Generating explanations for model decisions is a step towards accountable AI, allowing for the scrutiny of model behavior and the identification of ethical issues such as biases or privacy concerns. It supports compliance with regulations and ethical guidelines that demand transparency and explainability in AI systems.
    \item \textbf{Improving User Experience:} For end-users, especially those without technical expertise, language-based explanations demystify AI operations, making LLMs more approachable and their outputs more trustworthy. This can significantly improve user experience and satisfaction in applications ranging from customer service chatbots to AI-assisted decision-making tools.
\end{itemize}

\subsection{Techniques and Considerations}
\begin{itemize}
    \item \textbf{Self-Explanation Models:} Some LLMs are designed or fine-tuned to generate explanations for their own predictions or decisions as part of their output. This self-explanation capability requires careful training and validation to ensure that the explanations are accurate, relevant, and genuinely reflective of the model's decision-making process.
    \item \textbf{Dedicated Explanation Models:} Alternatively, a separate model can be trained to generate explanations for the outputs of an LLM. This approach allows for flexibility and specialization in explanation generation but requires careful coordination to ensure that the explanation model accurately captures and communicates the reasoning of the primary LLM.
    \item \textbf{Evaluation of Explanation Quality:} Assessing the quality of language-based explanations involves evaluating their accuracy (do they correctly reflect the model's reasoning?), completeness (do they cover all relevant aspects of the decision?), and comprehensibility (are they easily understood by humans?). Developing metrics and methodologies for this evaluation is an ongoing challenge in the field.
    \item \textbf{Bias and Misinterpretation:} There's a risk that language-based explanations might introduce or perpetuate biases, or be misinterpreted by users. Ensuring that explanations are clear, unbiased, and accurately represent the model's operations is crucial.
\end{itemize}

\subsection{Challenges}
\begin{itemize}
    \item \textbf{Complexity of Generating High-Quality Explanations:} Producing explanations that are both accurate and easily understandable by non-experts is challenging, especially for complex decisions or abstract concepts.
    \item \textbf{Scalability:} Generating tailored explanations for every output can be computationally intensive, particularly for large-scale or real-time applications.
    \item \textbf{Alignment with Human Reasoning:} Ensuring that machine-generated explanations align with human reasoning and expectations requires deep understanding of both the domain and human communication patterns.
\end{itemize}

Language-Based Explanations serve as a vital tool for making LLMs more interpretable, accountable, and user-friendly. By articulating the reasoning behind their outputs in natural language, LLMs can achieve greater transparency, fostering trust and enabling more effective human-machine collaboration. Developing effective strategies for generating and evaluating these explanations remains a key focus for advancing the field of AI interpretability and ethics.

\section{Embedding Space Analysis}
Embedding Space Analysis is an essential method for delving into the high-dimensional vector spaces (embeddings) utilized by Large Language Models (LLMs) to represent linguistic elements such as words and phrases. This analysis sheds light on the semantic and syntactic relationships encoded within these embeddings, offering valuable insights into the models' language processing and representation capabilities.

Liu (2019) delves into latent space cartography, a pioneering approach to mapping semantic dimensions within vector space embeddings, which holds significant implications for understanding the intricate semantic and syntactic interplay in LLMs. Saul (2001) introduces locally linear embedding (LLE), a dimensionality reduction algorithm with potential applications in analyzing LLM embedding spaces, suggesting a pathway to uncover the underlying structures within these complex models. Further, Almeida (2019) and Ruder (2017) offer thorough surveys on word embeddings, a foundational component of LLMs' vector spaces, providing insights into the construction and cross-lingual evaluation of word embeddings. These contributions collectively underscore the importance of Embedding Space Analysis in unpacking the nuanced ways LLMs understand and represent language, highlighting the technique's role in advancing our grasp of artificial linguistic intelligence.

\subsection{Application in LLM Evaluation}
\begin{itemize}
    \item \textbf{Discovering Semantic Relationships:} Embedding space analysis allows for the exploration of semantic relationships encoded by the LLM. By examining the distances and directions between vectors, researchers can identify clusters of related words or phrases, uncover synonyms and antonyms, and even detect more complex relationships like analogies.
    \item \textbf{Understanding Model Generalization:} The way embeddings are organized within the vector space can also offer clues about the model's ability to generalize across different contexts. A well-organized embedding space, where similar concepts are grouped together in a consistent manner, suggests that the model has a robust understanding of the underlying language structure.
    \item \textbf{Evaluating Contextual Understanding:} Modern LLMs, especially those based on Transformer architectures, generate context-dependent embeddings. Analyzing these context-specific embeddings can reveal how the model's representation of a word changes with its context, highlighting the model's capacity for nuanced language understanding.
    \item \textbf{Bias Detection:} Embedding spaces can inadvertently capture and amplify biases present in the training data. By analyzing embeddings, researchers can detect biases in how concepts are represented and related, which is crucial for developing more fair and unbiased models.
\end{itemize}

\subsection{Techniques and Considerations}
\begin{itemize}
    \item \textbf{Dimensionality Reduction:} Given the high-dimensional nature of embeddings, dimensionality reduction techniques (such as t-SNE or PCA) are often employed to visualize the embedding space in two or three dimensions. This visualization can make patterns and relationships more accessible and interpretable.
    \item \textbf{Cosine Similarity Analysis:} Cosine similarity is a common measure used to assess the similarity between two vectors in the embedding space. It allows for the quantitative comparison of semantic similarity between words or phrases, facilitating the systematic exploration of linguistic relationships.
    \item \textbf{Cluster Analysis:} Clustering algorithms can identify groups of similar embeddings, helping to uncover underlying structures or themes in the data. This analysis can highlight how the model categorizes concepts and whether these categorizations align with human understanding.
    \item \textbf{Probing Tasks:} Probing tasks are designed to directly test specific properties of embeddings, such as grammatical tense, number, or entity type. By evaluating the model's performance on these tasks, researchers can assess the depth and specificity of the linguistic information captured by the embeddings.
\end{itemize}

\subsection{Challenges}
\begin{itemize}
    \item \textbf{Interpretability:} While embedding space analysis can reveal complex patterns, interpreting these patterns and relating them back to model behavior or linguistic theory can be challenging. It requires a nuanced understanding of both the model architecture and the linguistic phenomena being investigated.
    \item \textbf{High-Dimensional Complexity:} The high-dimensional nature of embeddings means that much of the structure and information in the embedding space can be lost or obscured when using dimensionality reduction techniques for visualization.
    \item \textbf{Contextual Embeddings:} For models that generate context-dependent embeddings, the analysis becomes more complex, as the representation of a word or phrase can vary significantly across different contexts. This variability can make it harder to draw general conclusions about the model's linguistic understanding.
\end{itemize}

Embedding Space Analysis provides a powerful window into the inner workings of LLMs, offering insights into how these models process, understand, and represent language. By carefully examining the structures and patterns within embedding spaces, researchers and developers can enhance their understanding of LLM capabilities, biases, and potential areas for improvement, contributing to the development of more sophisticated, fair, and transparent language models.

\section{Computational Efficiency and Resource Utilization of LLMs}
The evaluation of Large Language Models (LLMs) extends beyond their linguistic prowess to include critical assessments of computational efficiency and resource utilization. Key performance indicators such as memory usage, CPU/GPU utilization, and model size are essential for optimizing LLM operations.

Gao (2002) and Heafield (2013) both contribute to enhancing language model efficiency, with Gao underscoring the significance of pruning criteria and Heafield pioneering efficient algorithms for language modeling challenges. Chilkuri (2021) introduces the Legendre Memory Unit, a novel architecture that markedly decreases the memory and computation demands for language modeling. Zhang (2023) shifts the focus to the strategic importance of instruction tuning, as opposed to merely increasing model size, for improving zero-shot summarization capabilities in LLMs.

These contributions highlight the ongoing imperative for advancements in the computational efficiency and judicious resource use of LLMs, underscoring the balance between model performance and operational sustainability.

\subsection{Memory Usage}
\begin{itemize}
    \item \textbf{Peak Memory Consumption:} The maximum amount of RAM required by the model during training or inference. This metric is crucial for understanding the scalability of the model across different hardware environments.
    \item \textbf{Memory Bandwidth Utilization:} Measures how efficiently the model uses the available memory bandwidth. High bandwidth utilization can indicate optimized memory access patterns, crucial for high-performance computing environments.
\end{itemize}

\subsection{CPU/GPU Usage}
\begin{itemize}
    \item \textbf{CPU/GPU Utilization Percentage:} The proportion of CPU or GPU resources utilized during model operations. High utilization rates can indicate efficient use of hardware resources but may also signal potential bottlenecks if consistently at capacity.
    \item \textbf{FLOPS (Floating Point Operations Per Second):} A measure of the computational power used by the model. Higher FLOPS indicate more intensive computation, which can be a double-edged sword—indicating either complex model capabilities or inefficiencies in computation.
    \item \textbf{Inference Time:} The time it takes for the model to generate an output given an input. Faster inference times are preferred for real-time applications, reflecting efficient CPU/GPU usage.
\end{itemize}

\subsection{Size of the Model}
\begin{itemize}
    \item \textbf{Number of Parameters:} Reflects the complexity and potential capacity of the model. Larger models, with billions or even trillions of parameters, can capture more nuanced patterns but are more demanding in terms of storage and computation.
    \item \textbf{Model Storage Size:} The disk space required to store the model. This is directly influenced by the number of parameters and the precision of the weights (e.g., using 16-bit vs. 32-bit floating-point numbers).
    \item \textbf{Compression Ratio:} After model pruning or quantization, the compression ratio indicates the efficiency of reducing the model size without significantly impacting performance. Higher ratios suggest effective size reduction while maintaining model accuracy.
\end{itemize}

\subsection{Energy Consumption}
\begin{itemize}
    \item \textbf{Watts per Inference/Training Hour:} Measures the energy required to perform a single inference or for an hour of model training. Lower energy consumption is desirable for reducing operational costs and environmental impact.
\end{itemize}

\subsection{Scalability}
\begin{itemize}
    \item \textbf{Parallelization Efficiency:} Indicates how well the model training or inference scales across multiple CPUs or GPUs. High efficiency means that adding more hardware resources proportionally decreases training/inference time.
    \item \textbf{Batch Processing Capability:} The ability of the model to process data in batches efficiently, impacting throughput and latency. Larger batch sizes can improve throughput but may also increase memory and computational requirements.
\end{itemize}

Understanding and optimizing these performance metrics are crucial for deploying LLMs effectively, especially in resource-constrained environments or applications requiring high throughput and low latency.

\section{Human Evaluation of LLMs}
Human Evaluation stands as an indispensable method for appraising Large Language Models (LLMs), complementing automated metrics with the discernment of human judges. This process involves evaluators, ranging from experts to general audiences, scrutinizing the generated text's quality, relevance, coherence, and ethical dimensions. Such evaluations tap into the subtleties and complexities of language that automated systems might miss, emphasizing the importance of subjective judgment and contextual understanding.

Turchi (2013) and Manning (2020) both underscore the significance of human judgment in evaluating LLM outputs, highlighting the nuanced insights human evaluators bring to the table. Lee (2021) points out the necessity for establishing standardized practices in human evaluation to ensure consistency and reliability across assessments. Addressing this, An (2023) introduces L-Eval, a framework aimed at standardizing the evaluation of long-context language models. This framework proposes a comprehensive evaluation suite, advocating for the use of Length-Instruction-Enhanced (LIE) evaluation methods and the incorporation of LLM judges, thereby advancing the methodologies for human evaluation of LLMs.

\subsection{Understanding Human Evaluation}
\textbf{Concept:} Human evaluation relies on individuals assessing the outputs of LLMs based on criteria such as linguistic quality (grammar, syntax), relevance to a prompt, coherence of the text, creativity, and alignment with ethical standards. This can involve direct rating scales, comparative assessments, or qualitative feedback.\\
\textbf{Application:} Evaluators are typically presented with outputs from the LLM alongside tasks or prompts. They might also compare these outputs against a reference standard or across different models to gauge performance. The evaluation can be structured around specific tasks (e.g., translation, summarization) or more open-ended assessments of generative text.

\subsection{Application in Evaluating LLMs}
\begin{itemize}
    \item \textbf{Qualitative Insights:} Human evaluation captures the subtleties of language and communication that automated metrics might miss, such as cultural nuances, emotional tone, and implicit meanings. This can be particularly important in applications like storytelling, content creation, and sensitive communications.
    \item \textbf{Benchmarking Real-World Usability:} By assessing how well model-generated text meets human expectations and needs, evaluators can determine the model's readiness for real-world applications. This includes understanding user satisfaction and potential areas of improvement for better alignment with human users.
    \item \textbf{Identifying Ethical and Societal Impacts:} Human judges can evaluate text for biases, stereotypes, or potentially harmful content, providing insights into the ethical and societal implications of deploying LLMs at scale.
    \item \textbf{Enhancing Model Training and Development:} Feedback from human evaluation can guide further model training and refinement, especially in improving the model's handling of complex, nuanced, or culturally specific content.
\end{itemize}

\subsection{Challenges and Considerations}
\begin{itemize}
    \item \textbf{Subjectivity and Variability:} Human judgments can vary significantly between individuals, influenced by personal experiences, cultural backgrounds, and subjective preferences. Establishing consistent evaluation criteria and training evaluators can help mitigate this variability.
    \item \textbf{Scalability and Cost:} Human evaluation is resource-intensive, requiring significant time and effort from skilled individuals. Balancing thoroughness with practical constraints is a key challenge, especially for large-scale models and datasets.
    \item \textbf{Bias and Fairness:} Evaluators' biases can influence their assessments, potentially introducing subjective biases into the evaluation process. Diverse and representative panels of evaluators can help address this concern.
    \item \textbf{Integration with Automated Metrics:} For a comprehensive evaluation, human assessments should be integrated with automated metrics, balancing the depth of human insight with the scalability and consistency of automated evaluations.
\end{itemize}

\section{Conclusion and Future Work}
Our investigation into evaluation methodologies for Large Language Models (LLMs) underscores the critical need for transparent, understandable, and ethical AI systems, particularly within educational contexts such as the AI for Education Project (AI4ED) at Northeastern University. This initiative exemplifies the potential of AI to revolutionize educational practices by providing adaptive and personalized learning experiences.

Key points from our study include:

\begin{itemize}
    \item \textbf{LLMMaps:} This innovative visualization technique offers a nuanced evaluation of LLMs across various NLP subfields, highlighting performance strengths and areas needing improvement, with a focus on reducing hallucinations.
    \item \textbf{Benchmarking and Leaderboards:} These tools provide systematic assessments of LLM performance on extensive Q\&A datasets, promoting competition and progress in model development.
    \item \textbf{Stratified Analysis:} This method dissects LLM performance into distinct layers or strata, enabling detailed insights into model strengths and weaknesses across different knowledge subfields.
    \item \textbf{Visualization of Bloom’s Taxonomy:} This approach visualizes model performance across cognitive skill levels, aiding in the assessment of LLM capabilities in handling tasks of varying complexity.
    \item \textbf{Hallucination Score:} This metric measures the frequency and severity of hallucinations in LLM outputs, guiding efforts to mitigate inaccuracies and enhance model reliability.
    \item \textbf{Knowledge Stratification Strategy:} This method organizes Q\&A datasets into hierarchical knowledge structures, facilitating detailed analysis of LLM performance across various levels of complexity and domain specificity.
    \item \textbf{Utilization of Machine Learning Models for Hierarchy Generation:} This approach employs machine learning algorithms to classify and arrange questions into coherent hierarchies, enhancing systematic LLM evaluation.
    \item \textbf{Sensitivity Analysis:} This technique assesses LLM responsiveness to input variations, revealing insights into model robustness and decision-making patterns.
    \item \textbf{Feature Importance Methods:} These methods pinpoint crucial input features influencing model outputs, enhancing transparency and guiding model improvement efforts.
    \item \textbf{Shapley Values:} Derived from cooperative game theory, Shapley values provide a fair and robust measure of individual input feature contributions, offering deep insights into LLM decision-making processes.
    \item \textbf{Attention Visualization:} This technique elucidates how LLMs allocate importance to various input elements, enhancing understanding of model focus and decision-making strategies.
    \item \textbf{Counterfactual Explanations:} This method explores how slight input modifications affect model outputs, revealing underlying causal mechanisms and enhancing transparency.
    \item \textbf{Language-Based Explanations:} These explanations translate LLM decision-making processes into natural language, making model outputs more understandable and accessible.
    \item \textbf{Embedding Space Analysis:} This method examines high-dimensional vector spaces used by LLMs to represent linguistic elements, offering insights into semantic and syntactic relationships.
    \item \textbf{Computational Efficiency and Resource Utilization:} Key performance indicators such as memory usage, CPU/GPU utilization, and model size are crucial for optimizing LLM operations.
    \item \textbf{Human Evaluation:} Involving human judges to assess LLM outputs provides qualitative insights that complement automated metrics, capturing nuances and ethical considerations.
\end{itemize}

Future work should prioritize the evaluation of these methodologies within AI4ED, focusing on their applicability and effectiveness in educational settings. Additionally, there is a pressing need for further research on visualizing these evaluation techniques in a manner that is accessible to students, administrators, and faculty alike. By bridging the gap between complex AI technologies and their practical application in education, we can foster a deeper understanding and integration of AI tools in enhancing learning outcomes, aligning with Northeastern University's mission to lead in innovative educational methodologies.

\end{document}